\documentclass[letterpaper, 10 pt, conference]{ieeeconf}  % Comment this line out if you need a4paper

\IEEEoverridecommandlockouts                              % This command is only needed if 
\usepackage{graphicx, epstopdf, subfig, amsmath, diagbox,multirow,algorithm,algorithmic,url}
\usepackage{float}
\usepackage{stfloats}
\usepackage{color}

\title{\LARGE \bf
Unsupervised Trajectory Segmentation and Promoting \\ of Multi-Modal Surgical Demonstrations
}

\author{Zhenzhou Shao$^{1}$, Hongfa Zhao$^{1}$, Jiexin Xie$^{1}$, Ying Qu$^{2*}$, Yong Guan$^{1}$  and Jindong Tan$^{2}$ % <-this % stops a space
	\thanks{*Corresponding author.}% <-this % stops a space
	\thanks{$^{1}$Zhenzhou Shao, Hongfa Zhao, Jiexin Xie and Yong Guan are with the College of Information Engineering, Beijing Advanced Innovation Center for Imaging Technology and Beijing Key Laboratory of Light Industrial Robot and Safety Verification, Capital Normal University, Beijing, 100048, China
		{\tt\small \{zshao,hfzhao,2171002039,guanyong\}@cnu.edu.cn}}
	\thanks{$^{2}$Ying Qu and Jindong Tan are with the Engineering College, The University of Tennessee, Knoxville, TN, 37996, USA
		{\tt\small \{yqu3,tan\}@utk.edu}}%
}

\begin{document}

\maketitle
\thispagestyle{empty}
\pagestyle{empty}

%%%%%%%%%%%%%%%%%%%%%%%%%%%%%%%%%%%%%%%%%%%%%%%%%%%%%%%%%%%%%%%%%%%%%%%%%%%%%%%%
\begin{abstract}
To improve the efficiency of surgical trajectory segmentation for robot learning in robot-assisted minimally invasive surgery, this paper presents a fast unsupervised method using video and kinematic data, followed by a promoting procedure to address the over-segmentation issue. Unsupervised deep learning network, stacking convolutional auto-encoder, is employed to extract more discriminative features from videos in an effective way. To further improve the accuracy of segmentation, on one hand, wavelet transform is used to filter out the noises existed in the features from video and kinematic data. On the other hand, the segmentation result is promoted by identifying the adjacent segments with no state transition based on the predefined similarity measurements. Extensive experiments on a public dataset JIGSAWS show that our method achieves much higher accuracy of segmentation than state-of-the-art methods in the shorter time. 
\end{abstract}

%%%%%%%%%%%%%%%%%%%%%%%%%%%%%%%%%%%%%%%%%%%%%%%%%%%%%%%%%%%%%%%%%%%%%%%%%%%%%%%%
\section{INTRODUCTION}

Surgical trajectory segmentation is a fundamental problem in the field of robot-assisted minimally invasive surgery (RMIS). It can be applied to several applications, such as demonstration learning \cite{guha2013minimalist}, skill assessment \cite{reiley2008automatic}, complex task automation \cite{shamaei2015paced} and so forth. Each surgical procedure is usually represented by synchronized video and kinematic recordings, and can be decomposed into several meaningful sub-trajectories. Since the segments are atomic with less complexity, lower variance and easier to eliminate outliers, the capability of further robot learning and assessment can be improved. However, it is a challenging task to segment the surgical trajectory accurately and rapidly. Even an identical surgical procedure can vary remarkably in the spatial and temporal domains due to the skill difference among surgeons. Moreover, the trajectory is susceptible to the random noise.

%such as demonstration learning \cite{reiley2010motion,guha2013minimalist}, skill assessment \cite{mackenzie2001hierarchical,reiley2008automatic}, complex task automation \cite{murali2015learning,shamaei2015paced}

Traditional solutions usually transfer the surgical trajectory segmentation to a clustering problem, and are mainly divided into two categories: supervised and unsupervised methods. As the supervised methods, Linear Discriminate Analysis (LDA) \cite{lin2005automatic}, Hidden Markov Models (HMMs) \cite{reiley2008automatic}, Descriptive Curve Coding (DCC) \cite{ahmidi2013string}, and Conditional Random Field (CRF) \cite{tao2013surgical} are proposed. However, the supervised method is time-consuming because of the manual annotations of experts for training dataset. Thus, unsupervised methods have drawn more attention in recent years. Some unsupervised methods based on Gaussian Mixture Model (GMM) and Dirichlet Processes (DP) are proposed \cite{lee2015autonomous,krishnan2017transition}. Although GMM and DP based methods can get rid of the manual annotations, the room to improve the accuracy of surgical trajectory segmentation remains since only the kinematic data is taken into account. Recently, video data are involved by using a deep learning based method, since traditional pattern recognition based feature extraction methods can't model the variations among surgeon's videos well. A. Murali \textit{et al.} \cite{murali2016tsc} employ VGGNet to extract features from video followed by Transition State Clustering (TSC) for task-level segmentation using both kinematic and video data. Although the involvement of video source enables the higher accuracy of segmentation, the feature extraction from videos is time-consuming and easily leads to over-segmentation.

This paper focuses on the unsupervised surgical trajectory segmentation by means of both video and kinematic data in this paper. There are challenges to find consistent segments from the varying and noising recordings from surgeons with different skills for a specific task. First, although the video is capable of improving the performance of segmentation, it is challenging to extract the distinguishing features in an efficient way. In addition, random noise has to be considered due to the difference of surgeons’ skill. Second, state-of-the-art methods generally suffer from the over-segmentation issue. We need to provide an effective way to identify the adjacent segments with no state transition.

\begin{figure}[pb]
	\vspace{-10mm}  %调整图片与上文的垂直距离
	\setlength{\abovecaptionskip}{-3mm}   %调整图片标题与图距离
	\setlength{\belowcaptionskip}{-0mm}   %调整图片标题与下文距离
	\centering
	\includegraphics[scale = 0.5]{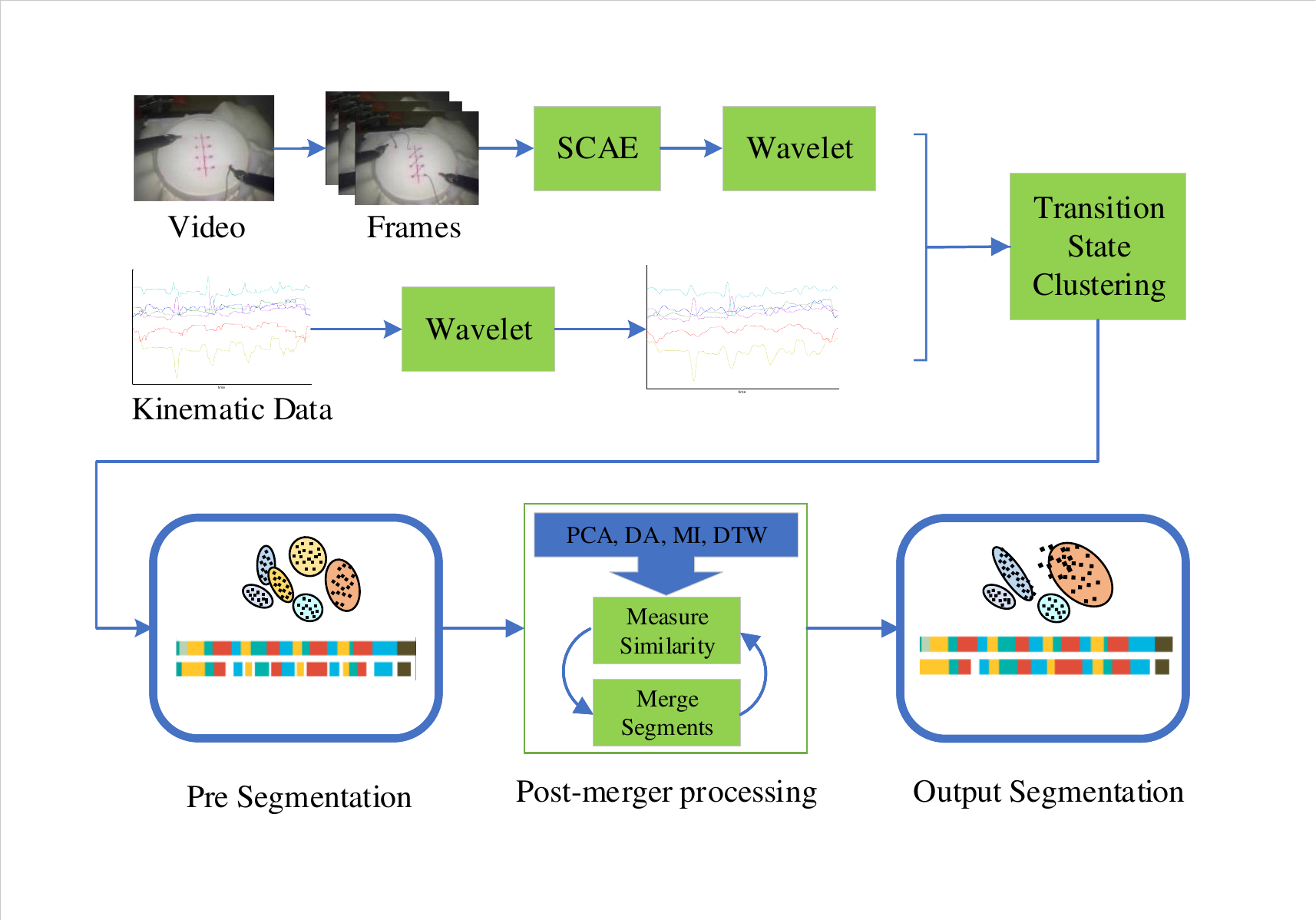}
	\caption{Illustration of the suturing trajectory segmentation with promoting procedure using video and kinematic data.}
	\label{fig:intro} %% label for first subfigure
\end{figure}

As shown in Fig. \ref{fig:intro}, a fast unsupervised method for surgical trajectory segmentation is proposed using the video and kinematic data. In particular, a promoting procedure is presented to alleviate the over-segmentation issue. First, a compact but effective unsupervised learning network called stacking convolutional auto-encoder (SCAE) is employed to speed up the feature extraction of video. Wavelet transform is then used to filter the features from videos and kinematic data for the further clustering based on TSC. We refer the proposed segmentation method as TSC-SCAE for abbreviation. Finally, the segmentation result is promoted by merging the clusters according to four similarity measurements called PMDD based on  principal component analysis, mutual information, data average and dynamic time warping, respectively.

\section{UNSUPERVISED TRAJECTORY SEGMENTATION BASED on TSC-SCAE}
\subsection{Visual Feature Extraction Using SCAE}

\begin{figure*}[bp] 
	\vspace{-5mm}  %调整图片与上文的垂直距离
	\setlength{\abovecaptionskip}{-0mm}   %调整图片标题与图距离
	\setlength{\belowcaptionskip}{-0mm}   %调整图片标题与下文距离
	\centering
	\includegraphics[scale = 0.8]{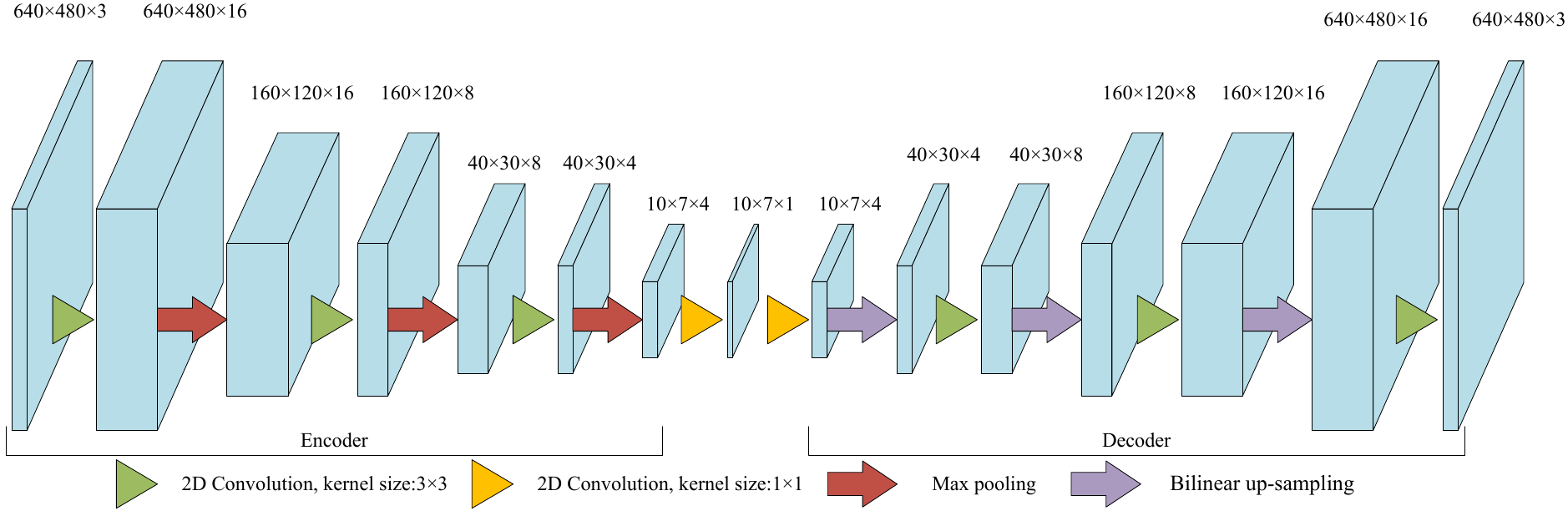} %0.8
	\caption{SCAE network for visual feature extraction.}%  \vspace{-5mm}
	\label{fig:scae} %% label for first subfigure
\end{figure*}

Stacked Convolutional Auto-Encoder (SCAE) \cite{masci2011stacked} is an unsupervised feature extractor which is well compatible to high-dimensional input. It is much faster than other methods such as TSC-VGG and TSC-SIFT because of the simple neural network and unsupervised method. SCAE has more advantages in image processing as it can preserve the spatial relationship between pixels. The SCAE network for visual feature extraction is shown in Fig. \ref{fig:scae}, and the corresponding configuration is summarized in TABLE \ref{tab:scae}.

Fig. \ref{fig:scae} illustrates that the basic structure of encoder consists of convolutional layer and pooling layer. The input feature maps (for the first layer, it is the original image $I$) are convolved with a convolution layer to transfer the information to subsequent layers with the spatial relationship between pixels preserved. These feature maps then pass through a max-pooling layer to reduce the feature map size. After several above conv-pooling layers, a low dimension feature map can get from the encoder.

As shown in Fig. \ref{fig:scae}, the task of the decoder with the similar topology with the encoder is to reconstruct the encoding result to get the implied image information. Therefore, we need to up-sample the encoding result to recover the feature maps. To prevent the checkerboard effect caused by traditional transposed convolution, we use bilinear interpolation to do up-sampling before each convolutional layer. For further reduction of feature dimension, we employ two convolutional layers with the kernel size of $1\times 1$ after the last layer of the encoder and before the first layer of decoder respectively.

Adam optimization algorithm \cite{kingma2014adam} is employed to minimize the MSE (mean-square error) based loss function, which can estimate the similarity between the reconstructed image ${{\hat{I}}}$ of decoder output and the original image $I$ input to encoder. After the network training, a model (i.e., the weights of each layer) for image encoding and reconstructing can be got from the network. In the phase of feature extraction, we exclusively load the model's encoder part to extract the features of each frame in the surgical video.

\begin{table}[h]
	\vspace{-3mm}  %调整图片与上文的垂直距离
	\centering
	\caption{Configuration of SCAE network.} 
	\label{tab:scae}
	\setlength{\tabcolsep}{5.1pt}
	\begin{tabular}{ccccc}
		\hline
		&         Type         & Patch Size & Stride &       Output Size        \\ \hline
		&     convolution      & 3$\times$3 &   1    & 640$\times$480$\times$16 \\
		&       max pooling       & 4$\times$4 &   4    & 160$\times$120$\times$16 \\
		&     convolution      & 3$\times$3 &   1    & 160$\times$120$\times$8  \\
		Encoder &       max pooling       & 4$\times$4 &   4    &  40$\times$30$\times$8   \\
		&     convolution      & 3$\times$3 &   1    &  40$\times$30$\times$4   \\
		&       max pooling       & 4$\times$4 &   4    &   10$\times$7$\times$4   \\
		&     convolution      & 1$\times$1 &   1    &   10$\times$7$\times$1   \\ \hline
		&     convolution      & 1$\times$1 &   1    &   10$\times$7$\times$4   \\
		& bilinear up-sampling &            &        &  40$\times$30$\times$4   \\
		&     convolution      & 3$\times$3 &   1    &  40$\times$30$\times$8   \\
		Decoder & bilinear up-sampling &            &        & 160$\times$120$\times$8  \\
		&     convolution      & 3$\times$3 &   1    & 160$\times$120$\times$16 \\
		& bilinear up-sampling &            &        & 640$\times$480$\times$16 \\
		&     convolution      & 3$\times$3 &   1    & 640$\times$480$\times$3  \\ \hline
	\end{tabular}
	\vspace{-3mm}  %调整图片与上文的垂直距离
\end{table}

\subsection{Denoising Based on Wavelet Transform}

After the feature extraction from the demonstration video, the visual and kinematic features are then feed to nonparametric mixture model for clustering. However, we find that these features usually suffer from the random noise. To get rid of it, wavelet transform based filter is employed due to its ability of multi-scale filtering and a low-pass filter is designed. 

%By the use of wavelet transform, we divide operation trajectory into high frequency and low frequency domains. The high frequency part reflects the trajectory differences in the details, while the low frequency one can be seen as a smooth approximation to the original data. Medical trajectory segmentation mainly refers to low frequency section, as a consequence, high frequency section may become a hindrance under certain circumstances. Low pass filter, which can effectively avoid the small-scale noises (always high frequency section) caused by surgeon's remote operation to some extent. The video sources follow the same rules.

\begin{figure}[h]
	\vspace{0mm}  %调整图片与上文的垂直距离
	\setlength{\abovecaptionskip}{0mm}   %调整图片标题与图距离
	\setlength{\belowcaptionskip}{-3mm}   %调整图片标题与下文距离
	\centering
	\includegraphics[scale = 0.5]{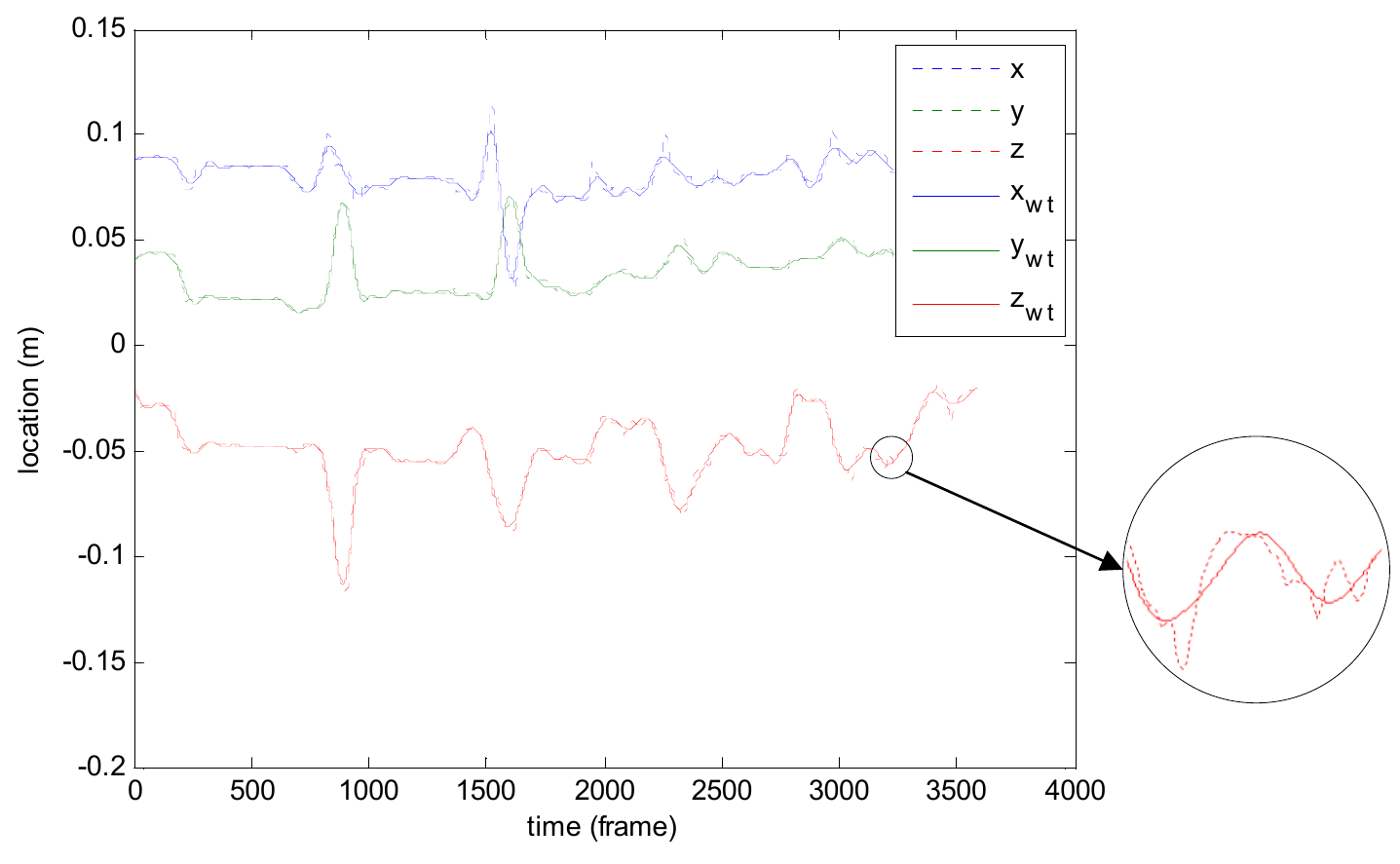}
	\caption{Comparison of kinematic features  before and after filtering based on wavelet transform: The unit of vertical Y-axis is meter (m) and the horizontal X-axis is frame (30 fps), $x$, $y$, $z$ are input data (which represent a spatial location comprehensively) while $x_{wt}$, $y_{wt}$, $z_{wt}$ are the corresponding denoising results.}%  \vspace{-5mm}
	\label{fig:wavelet_k} %% label for first subfigure
\end{figure}

\begin{figure}[h]
	\vspace{-0mm}  %调整图片与上文的垂直距离
	\setlength{\abovecaptionskip}{0mm}   %调整图片标题与图距离
	\setlength{\belowcaptionskip}{-5mm}   %调整图片标题与下文距离
	\centering
	\includegraphics[scale = 0.5]{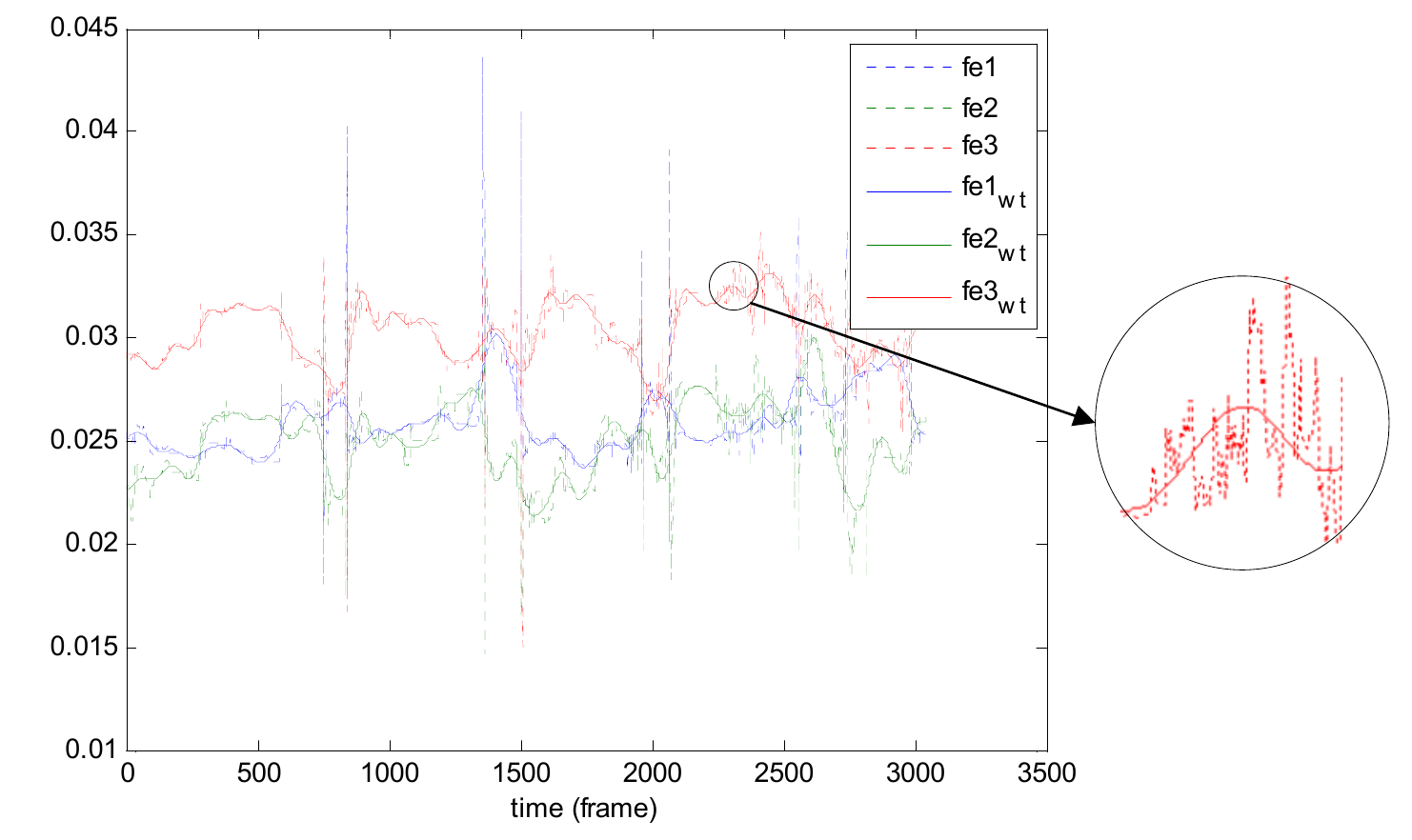}
	\caption{Comparison of visual features before and after filtering based on wavelet transform: The horizontal X-axis is frame (30 fps), Y-axis denotes the value of visual feature. 
		%The visual features are in 70 dimensions in the experiment, only 3 dimensions of them are selected for visualization.
	}
	\label{fig:wavelet_v} %% label for first subfigure
\end{figure}

In this paper, we process the kinematic data and visual features with db10 wavelet, and a 5-level wavelet decomposition for denoising is performed. Fig. \ref{fig:wavelet_k} and Fig. \ref{fig:wavelet_v} demonstrate the comparison of kinematic and visual features before and after filtering based on wavelet transform.

After the filtering, visual and kinematic features then feed to a nonparametric mixture model to segment surgical trajectory. Considering the clustering performance, Transition State Clustering (TSC) \cite{krishnan2017transition} is adopted in this paper.

\section{SEGMENTATION PROMOTING BASED on PMDD}

Most unsupervised trajectory segmentation methods usually have the problem of over-segmentation. To correct the wrongly segmented sub-trajectories that belong to the same cluster, a criterion is required to evaluate the similarity between segments. Taking a deep insight into the same sub-trajectory, they have a few implicit and explicit associations. Besides the similarity in spatial and temporal space, inner structure, variation node and moving trend are also the important factors. Taking these factors into consideration, we proposed a promoting algorithm based on PMDD consisting of four similarity measurements based on Principal Component Analysis (PCA), Mutual Information (MI), Data Average (DA) and Dynamic Time Warping (DTW).

\textbf{Similarity measurement based on PCA:} W. Krzanowski \textit{et al.} \cite{krzanowski1979between} show that the PCA can be used to measure the similarity between segments. PCA mainly determine the internal link and structure between the segments. Considering two segments $S_{a}$ and $S_{b}$, PCA could find several principle components of $S_{a}$ and $S_{b}$, which make up a subspace representing the main information of $S_{a}$ and $S_{b}$. The smaller subspace angle between $S_{a}$ and $S_{b}$ means the greater internal consistency between them. Thus, Similarity measurement based on PCA is defined by the angles between their subspaces comprised of principle components:
\begin{equation}
\setlength\abovedisplayskip{3pt}
\setlength\belowdisplayskip{3pt}
\label{eq:PCA}
{{SM}_{PCA}}\left( {{S}_{a}},{{S}_{b}} \right)=\frac{1}{q}\sum\limits_{i=1}^{q}{\sum\limits_{j=1}^{q}{\theta \left( i,j \right)}},
\end{equation}
where $q$ is the number of principle components.

\textbf{Similarity measurement based on MI:} The surgery is a continuous process, the data change of the segments in same surgery sub-process is similar. Entropy can be interpreted as a measurement of the uncertainty of the particular variables. Therefore, MI is a good similarity measurement for variation degree between two segments, which is obtained by subtracting the joint entropy $H\left( {{S}_{a}},{{S}_{b}} \right)$ from the entropy $H\left( {{S}_{a}} \right)$ and $H\left( {{S}_{b}} \right)$ of both segments:
\begin{equation}
\setlength\abovedisplayskip{3pt}
\setlength\belowdisplayskip{3pt}
\label{eq:MI}
{{SM}_{MI}}\left( {{S}_{a}},{{S}_{b}} \right)=H\left( {{S}_{a}} \right)+H\left( {{S}_{b}} \right) - H\left( {{S}_{a}},{{S}_{b}} \right),
\end{equation}

\textbf{Similarity measurement based on DA:} DA mainly reflects the spatial characteristic. During a sub-process of surgery, the trajectory in a short time interval is similar in the spatial space. Therefore, the distance between the centers of segments in spatial space is taken into account, as written as follows:
\begin{equation}
\setlength\abovedisplayskip{3pt}
\setlength\belowdisplayskip{3pt}
\label{eq:DA}
{{SM}_{DA}}\left( {{S}_{a}},{{S}_{b}} \right)=\left\| {{\mu}_{a}}-{{\mu}_{b}} \right\|,
\end{equation}
where $\mu_{a}$ and $\mu_{b}$ are mean vectors of segments $S_{a}$ and $S_{b}$.

\textbf{Similarity measurement based on DTW:} Due to the difference of surgeons' skill, the same action may show different sub-trajectories. One typical is the same behavior of different performance in temporal domain. The key issue of DTW is warping curve. Here, we take cumulative distance $\gamma \left( i,j \right)$ to calculate the best warping path while measure DTW similarity \cite{Berndt1996Finding}.
\begin{equation}
\setlength\abovedisplayskip{3pt}
\setlength\belowdisplayskip{3pt}
\label{eq:DTWK}
SM_{DTW}\left( {{S}_{a}},{{S}_{b}} \right)=\min \left( \sqrt{\sum\nolimits_{k=1}^{K}{{{w}_{k}}}}/K \right),
\end{equation}
where $w_{k}$ is the $k-th$ element of warping path, $K$ is the compensation parameter that can be identified by cumulative distance.
\begin{equation}
\setlength\abovedisplayskip{3pt}
\setlength\belowdisplayskip{3pt}
\label{eq:DTW}
\gamma \left( i,j \right)=d\left( {{q}_{i}},{{c}_{j}} \right)+\min \left\{ 
\begin{matrix} \gamma \left( i-1,j-1 \right) \\
\gamma \left( i-1,j \right) \\
\gamma \left( i,j-1 \right) \\
\end{matrix} \right.,
\end{equation}
where $d\left( {{q}_{i}},{{c}_{j}} \right)$ is the Euclidean distance between point $q_{i}$ and $c_{j}$.

All above four similarity measurements are in different dimensions. Thus, the normalization is required to obtain the final measure. For $SM_{PCA}$, $SM_{DA}$, $SM_{DTW}$, the smaller the value is, the more similar the two segments are. We perform the normalization of them using Eq. \eqref{eq:PDD_normal}, and the normalization for $S_{MI}$ is perform using Eq. \eqref{eq:M_normal}. After that, the final similarity $O$ can be calculated by Eq. \eqref{eq:O_normal}.
\begin{equation}
	\setlength\abovedisplayskip{3pt}
	\setlength\belowdisplayskip{3pt}
	\label{eq:PDD_normal}
	\resizebox{.85\hsize}{!}{$
		Y=\left\{ \begin{matrix}
		0,SM\ge mean\left( SM \right)  \\
		\frac{mean\left( SM \right)-SM}{mean\left( SM \right)-\min \left( SM \right)},SM<mean\left( SM \right)  \\
		\end{matrix} \right.,
		$}
\end{equation}
\begin{equation}
\setlength\abovedisplayskip{3pt}
\setlength\belowdisplayskip{3pt}
\label{eq:M_normal}
\resizebox{.85\hsize}{!}{$
Y=\left\{ \begin{matrix}
\frac{SM-mean\left( SM \right)}{\max \left( SM \right)-mean\left( SM \right)},SM>mean\left( SM \right)  \\
0,SM\le mean\left( SM \right)  \\
\end{matrix} \right.,
$}
\end{equation}
\begin{equation}
\setlength\abovedisplayskip{3pt}
\setlength\belowdisplayskip{3pt}
\label{eq:O_normal}
\resizebox{.85\hsize}{!}{$
{{O}_{a,b}}={{\left[ \frac{{{\left( {{Y}_{PCA}} \right)}^{2}}+{{\left( {{Y}_{DA}} \right)}^{2}}+{{\left( {{Y}_{DTW}} \right)}^{2}}+{{\left( {{Y}_{MI}} \right)}^{2}}}{4} \right]}^{{\scriptstyle{}^{1}/{}_{2}}}},
$}
\end{equation}

Then, according to final similarity measurements, segments that have high similarity can be merged iteratively. Considering the segmentation results $S=\left\{ {{S}_{i}},1\le i\le n \right\}$, the final similarity of each pair of two adjacent segments will be calculated by Eq. \eqref{eq:O_normal} in each iteration, and  then we obtain a set of results $O=\left\{ {{O}_{1,2}},{{O}_{2,3}},...,{{O}_{n-1,n}} \right\}$. Merge the pairs with the highest final similarity, and update comprehensive similarity $ O $, merge the most similar segments in the next iteration, until overall final similarity ${{O}_{\left( a,b \right)}}$ smaller than threshold $\tau$. The segmentation promoting algorithm is summarized in Algorithm \ref{alg:post_merger}.

\vspace{-2mm}  %调整图片与上文的垂直距离
\begin{algorithm}[h]
	\caption{Segmentation promoting based on PMDD.}  
	\label{alg:post_merger} 
	\textbf{Input:} Segments $S$, Threshold $\tau$.\\
	\begin{algorithmic}[1]
		\vspace{-5mm}  %调整图片与上文的垂直距离
		\WHILE {$O>\tau$}
		\FOR {$i=1:length(S)-1$}
		\STATE Calculate ${{O}_{\left( a,b \right)}}$ by Eq. \eqref{eq:O_normal}. 
		\ENDFOR
		\STATE $index\leftarrow \underset{i}{\mathop{\arg \max }}\,\left( {{O}_{i,i+1}} \right)$ 
		\STATE ${{S}_{index}}\leftarrow merge\left( {{S}_{index}},{{S}_{index+1}} \right)$
		\STATE $remove\left({S}_{index+1}\right)$
		\ENDWHILE
	\end{algorithmic} 
	\textbf{Output:} Post processed segments $S$. 
\end{algorithm}
\vspace{-2mm}  %调整图片与上文的垂直距离

\section{EXPERIMENTAL RESULTS}

In this section, two sets of experiments are conducted to verify the performance of proposed unsupervised segmentation algorithm for surgical trajectory. In the first experiment, TSC-SCAE is evaluated with respect to the accuracy and overall running time, compared with the classic clustering methods including GMM and TSC. The effects of different data sources and wavelet transform based filtering are analyzed quantitatively. Second, the promoting method of segmentation is verified by following different methods using the kinematic data alone and the combination of video and kinematic data, respectively.

The dataset JIGSAWS \cite{gao2014jhu} from Johns Hopkins University is used in the experiments, including data recordings and manual annotations. Data recordings consist of surgical video and kinematic data collected from Da Vinci Surgical System. The sampling frequency for both video and kinematics sources is 30Hz. The dataset contains three surgical tasks: Suturing (SU), Needle-Passing (NP) and Knot-Tying (KT), which are performed and annotated by 8 surgeons with different skill levels. The suturing and needle passing task are commonly used in literatures. In this paper, we adopt 11 demonstrations of these two tasks in the experiments, including the videos and kinematic data from 5 experts (E), 3 intermediates (I) and 3 novice (N). The kinematic data are in 38 dimensions, including position, angle velocity, angle of grasper, etc. All 11 videos of each task are used for SCAE model training and features extraction. The computational configuration used in the experiments is summarized in TABLE \ref{tab:configration}.

\begin{table}[h]
	\vspace{-3mm}  %调整图片与上文的垂直距离
	\centering
	\caption{Configuration used in the experiments.}
	\label{tab:configration}
	\begin{tabular}{cl}
		\hline
		Category &Specification \\ \hline
		Operating System	&Ubuntu \\
		CPU	&32 Intel Xeon E5-2620 v4 @ 2.10GHz \\
		GPU	&NVidia Tesla K40 \\
		CUDA Compute Capability & 3.5 \\
		CUDA Cores & 2880	\\
		RAM	&128GB \\
		Programming Language	&Python \\ \hline
	\end{tabular}
\vspace{-3mm}  %调整图片与上文的垂直距离
\end{table}

\subsection{Quantitative Analysis of TSC-SCAE}
\subsubsection{Accuracy Comparison}

In this section, the accuracy of TSC-SCAE is compared using Normalize Mutual Information (NMI), which indicates the transfer status similarity between a predictive clustering result $ A $ and the ground truth $ B $ (manual annotations), it can be calculated by
\begin{equation}
\setlength\abovedisplayskip{3pt}
\setlength\belowdisplayskip{3pt}
\label{eq:NMI}
NMI\left( A,B \right)=\frac{I\left( A,B \right)}{\sqrt{H\left( A \right)H\left( B \right)}},
\end{equation}
where $H\left( A \right)$ and $H\left( B \right)$ are the information entropies of  $ A $ and  $ B $, respectively. $I\left( A,B \right)$ is mutual information. The range of NMI is [0,1], where 0 means that there is no correlation between two clustering results, while 1 represents the results are completely related.

We compare the proposed method TSC-SCAE with state-of-the-art methods, including TSC\cite{krishnan2017transition}, GMM\cite{lee2015autonomous}, TSC-VGG, TSC-SIFT\cite{murali2016tsc}  and TSC-SCAE on the selected surgical demonstrations. According to the data source in the different methods, the experiments are divided into two categories: one use kinematics data alone and the other use both video and kinematic data. TABLE \ref{tab:NMI} shows NMI measurements of segmentation. We can see that our method TSC-SCAE achieves the best NMI among all trajectory segmentation tasks, it thanks to the use of video data and wavelet transform. Especially, using both video and kinematic data, the accuracy is improved by more than 2.6 times at most, compared with TSC-SIFT.

\begin{table}[h]
	\vspace{-3mm}  %调整图片与上文的垂直距离
	\centering
	\caption{NMI of segmentation for different methods. $ K $ stands for using
		kinematics data alone, $V\&K$ represents using both video and kinematics
		data, $ * $ denotes data is filtered by wavelet transform.}
	\label{tab:NMI}
	\setlength{\tabcolsep}{5.4pt}
	\begin{tabular}{c|ccc|ccc}
		\hline
		\multirow{2}{*}{\diagbox{Method}{NMI($\%$)}} & \multicolumn{3}{c|}{Needle Passing}              & \multicolumn{3}{c}{Suturing}                    \\ \cline{2-7} 
		& E              & E+I            & E+I+N          & E              & E+I            & E+I+N          \\ \hline
		TSC(K)          & 21.6          & 27.2          & 17.0          & 43.2          & 38.0          & 25.7          \\
		GMM(K)          & 53.3          & 51.2          & 45.8          & 45.2          & 43.4          & 41.0          \\
		%TSC-SCAE(K*)    & \textbf{25.8}          & \textbf{31.3}          & \textbf{28.2}          & \textbf{50.4}          & \textbf{46.5 }         &\textbf{ 37.4}          \\ 
		\hline
		TSC-VGG(V\&K)   & 62.9          & 64.7          & 69.3          & 58.6          & 64.0          & 66.5          \\
		TSC-SIFT(V\&K)  & 31.0          & 32.6          & 28.2          & 48.0          & 42.5          & 37.7          \\
		GMM-SCAE(V\&K)  & 59.3          & 57.4          & 58.7          & 57.5          & 52.5          & 51.4          \\
		TSC-SCAE(V\&K)  & 72.6          & 73.8          & 71.2          & 65.5          & 66.3          & 67.2          \\
		TSC-SCAE(V\&K*) & \textbf{79.1} & \textbf{77.7} & \textbf{74.7} & \textbf{67.9} & \textbf{67.5} & \textbf{68.5} \\ \hline
	\end{tabular}
	\vspace{-3mm}  %调整图片与上文的垂直距离
\end{table}

Overall, methods with both video and kinematic data are generally better than the ones using kinematics data alone. It is consistent with the results reported in literatures. The NMI of methods using kinematics data alone has a trend of decreasing with the growing proportion of non-expert (I $\&$ N) demonstrations. This phenomenon is very significant in the suturing task, it is mainly because of the complexity and non-regularity of suturing task. What’s more, demonstrations from experts are usually smoother and rapider than non-experts do. However, when considering both kinematics and video data, the phenomenon is obviously weakened. It proves that video data can help eliminate the influence of irregular trajectory from intermediates and novices and is an effective compensation to achieve the better surgical trajectory segmentation.

As aforementioned, random noise may cause the potential interference to the result of segmentation. To solve this problem, we perform a multi-scale smoothing processing to the dataset by using db10 wavelet to filter out the small-scale noise, which indirectly improve the segmentation accuracy. Compared with the experiments without filtering in needle-passing task, the NMI is increased by 3.5$\%$-6.5$\%$, the improvement is 1.2$\%$-3.4$\%$ in suturing task.

\subsubsection{Overall Running Time Comparison}

Another key indicator is overall running time, although surgery segmentation is not in strong real-time, the task also needs to be as fast as possible. Methods based on kinematics data alone, the running time is the cost of clustering and segmentation, while we need to add the time cost of video feature extraction for methods using visual and kinematic data (TSC-VGG, TSC-SIFT, etc.). For our method TSC-SCAE, the time cost is calculated in three parts, including visual feature extraction, wavelet transform based filtering and clustering segmentation.

% \usepackage{multirow}
\begin{table*}[pb]
	\vspace{-3mm}  %调整图片与上文的垂直距离
	\centering
	\caption{
		%Comparison of overall running time using different segmentation methods (unit: s), the dataset includes E (5 demonstrations from experts), E+I (3  and 3 demonstrations from experts and intermediates, respectively) and E+I+N (3, 3 and 3 demonstrations from experts, intermediates and novices). In the last column, 
		Comparison of overall running time using different segmentation methods (unit: s). FE stands for feature extraction, CS represents clustering segmentation and WT is wavelet transform, $ * $ denotes data is filtered by wavelet transform.}
	\label{tab:time}
	\setlength{\tabcolsep}{8.7pt}
	\begin{tabular}{c|ccc|ccc|c}
		\hline
		\multirow{2}{*}{\diagbox{Method}{Time(s)}} &               \multicolumn{3}{c|}{Needle Passing}               &                  \multicolumn{3}{c|}{Suturing}                   & \multirow{2}{*}{Elements} \\ \cline{2-7}
		&          E          &         E+I         &        E+I+N        &          E          &         E+I         &        E+I+N         &  \\ \hline
		TSC-K                    &         79          &         103         &         353         &         59          &         83          &         331          &            CS             \\
		GMM-K                    &        1.76         &        1.95         &        3.34         &        1.59         &        2.00         &         5.38         &            CS             \\
		TSC-VGG                   &      8120+394       &      9744+380       &     14616+1226      &      4935+322       &      5922+364       &      8884+1404       &           FE+CS           \\
		TSC-SIFT                  &      2127+440       &      3284+723       &      5019+2020      &      1941+404       &      3036+533       &      4633+2259       &           FE+CS           \\
		GMM-SCAE                  &      128+2.94       &      154+2.95       &      231+5.57       &      139+2.80       &      167+3.30       &       251+5.38       &           FE+CS           \\
		TSC-SCAE                 &        128+197       &        154+199        &       231+933        &        139+158        &        167+201        &        251+1012        &           FE+CS           \\
		TSC-SCAE*                  & \textbf{128+202+27} & \textbf{154+201+31} & \textbf{231+930+48} & \textbf{139+160+25} & \textbf{167+198+29} & \textbf{251+1008+47} &         FE+CS+WT          \\ \hline
	\end{tabular}
\vspace{-0mm}  %调整图片与上文的垂直距离
\end{table*}

\iffalse
\begin{figure}[h]
	\vspace{-3mm}  %调整图片与上文的垂直距离
	\setlength{\abovecaptionskip}{0mm}   %调整图片标题与图距离
	\setlength{\belowcaptionskip}{-5mm}   %调整图片标题与下文距离
	\centering
	\subfloat[Sturing Task]{
		\label{subfig:response-time-1} %% label for first subfigure
		\includegraphics[scale = 0.50]{figs/sturing_time}}
	\hspace{0.03in}
	\subfloat[Needle Passing Task]{% remember to revise the red rectangle.
		\label{subfig:response-time-2} %% label for second subfigure
		\includegraphics[scale = 0.50]{figs/needle_passing_time}}
	\caption{Overall running time of segmentation using different methods. CS stands for the time for clustering segmentation, FE for feature extraction and WT for  wavelet transform.}%  \vspace{-5mm}
	\label{fig:time} %% label for entire figure
\end{figure}
\fi

The running time in different steps is summarized in TABLE \ref{tab:time}. The segmentation methods based on both visual and kinematic features are about 10 times slower than the ones using kinematic data alone. It is mainly because of the time-consuming visual feature extraction. However, for the methods using both data sources, our method TSC-SCAE is almost 10 times faster than TSC-VGG and TSC-SIFT. The improvements of time efficiency is due to the high-efficiency unsupervised model for feature extraction of video data we employed. 

%As illustrated in Fig. \ref{fig:time}, 
%Through TSC-SCAE, we fuse video data and kinematics data rapidly and efficiently, make the segmentation result more accurately. Nevertheless, on an account of the nature characters of clustering segmentation algorithms, the result is always over-segmentation, this phenomenon is inevitable, needs post-merger processing.

\subsection{Evaluation of Segmentation Promoting}

Over-segmentation is a common problem of clustering segmentation algorithm. To prove the validity of the proposed promoting approach as the post-processing step, we apply it to the mainstream clustering segmentation algorithms, including GMM, TSC based methods. NMI is used to measure the similarity of transition status in the segmentation clustering method. But it is not based on transfer state to merge in the promoting stage. Therefore, we choose segmentation accuracy (seg-acc) as the evaluation matrix, which can measure the similarity between the segmentation result and ground truth intuitively and accurately. 

\begin{figure}[h]
	\vspace{-3mm}  %调整图片与上文的垂直距离
	\setlength{\abovecaptionskip}{1mm}   %调整图片标题与图距离
	\setlength{\belowcaptionskip}{-3mm}   %调整图片标题与下文距离
	\centering
	\includegraphics[scale = 0.46]{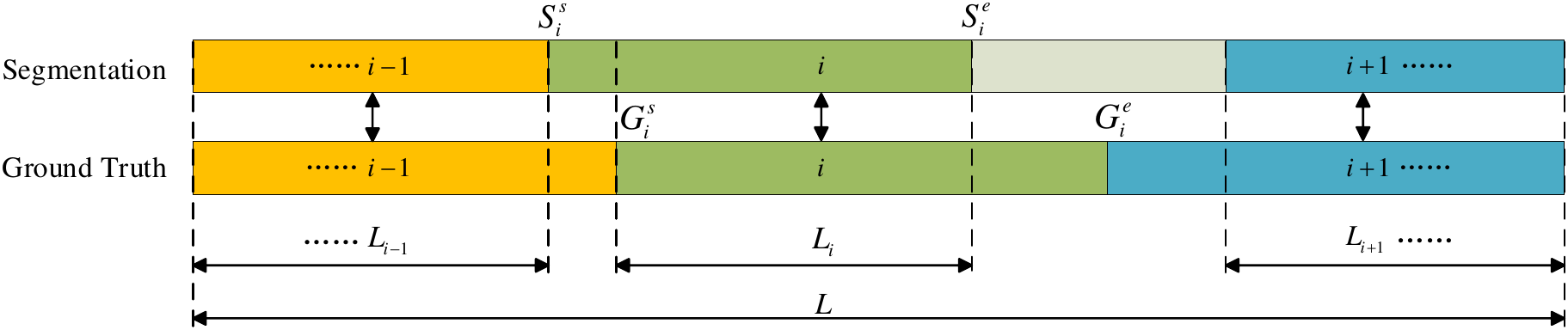}
	\caption{Segmentation accuracy of predicted segments. Where ‘$s$’ and ‘$e$’ represent start and end frame of segment, $L_{i}$ stands for the number of overlap frames between predicted segment $S_{i}$ and its corresponding ground truth $G_{i}$.}%  \vspace{-5mm}
	\label{fig:segacc} %% label for first subfigure
\end{figure}

%First, find the best mapping between ground truth and segments, second, calculate the similarity of each segment separately and then sum up them. 
The calculation of seg-acc can divided in two steps. In the first step, we match resultant segments to the ground truth by maximizing the number of overlap frames between predicted segments and ground truth \cite{wu2015watch}. In second step, it is true positive if the IOU (Intersection over Union) between the ground-truth segment $G_{i}$ and its corresponding resultant segments $S_{i}$ is more than a default threshold 40$\%$.  We calculate the accuracy of each segment separately and then sum up them.  Fig. \ref{fig:segacc} illustrates the calculation process and the seg-acc can be obtained using
\begin{equation}
\setlength\abovedisplayskip{3pt}
\setlength\belowdisplayskip{3pt}
\label{eq:segacc}
\resizebox{.85\hsize}{!}{$
seg\textrm{-}acc=\sum{\frac{{{L}_{i}}}{L}}=\sum{\frac{\left[ \min \left( S_{i}^{e},G_{i}^{e} \right)-\max \left( S_{i}^{s},G_{i}^{s} \right) \right]}{L}},
$}
\end{equation}
where ‘$S_{i}^{s},S_{i}^{e}$’ and ‘$G_{i}^{s},G_{i}^{e}$’ represent start and end frame of segment $S_{i}$ and $G_{i}$.

% \usepackage{multirow}
\begin{table*}[t]
	\vspace{-0mm}  %调整图片与上文的垂直距离
	\centering
	\caption{Segmentation accuracy before and after segmentation promoting.}
	\label{tab:segacc}
	\setlength{\tabcolsep}{7.9pt}
	\begin{tabular}{c|cccccc|cccccc}
		\hline
		\multirow{3}{*}{\diagbox{Method}{seg-acc\\ \\}} & \multicolumn{6}{c|}{Before Promoting }                                                               & \multicolumn{6}{c}{After Promoting }                                                                 \\
		 & \multicolumn{3}{c}{Needle Passing}               & \multicolumn{3}{c|}{Suturing}                    & \multicolumn{3}{c}{Needle Passing}               & \multicolumn{3}{c}{Suturing}                     \\  \cline{2-13}
		& E              & E+I            & E+I+N          & E              & E+I            & E+I+N          & E              & E+I            & E+I+N         & E              & E+I            & E+I+N          \\ \hline
		TSC-K             & 0.498          & 0.563          & 0.529          & 0.484          & 0.535          & 0.542          & 0.614          & 0.578          & 0.615          & 0.547          & 0.565          & 0.630          \\
		GMM               & 0.480          & 0.528          & 0.541          & 0.466          & 0.489          & 0.503          & 0.392          & 0.475          & 0.551          & 0.494          & 0.541          & 0.575          \\
		TSC-VGG           & 0.505          & 0.562          & 0.436          & 0.487          & 0.460          & 0.498          & 0.522          & 0.548          & 0.445          & 0.540          & 0.465          & 0.507          \\
		TSC-SIFT          & 0.546          & 0.561          & 0.510          & 0.442          & 0.513          & 0.493          & 0.592          & 0.582          & 0.590          & 0.521          & 0.589          & 0.593          \\
		TSC-SCAE          & \textbf{0.637} & \textbf{0.612} & \textbf{0.547} & \textbf{0.513} & \textbf{0.537} & \textbf{0.545} & \textbf{0.632} & \textbf{0.666} & \textbf{0.618} & \textbf{0.565} & \textbf{0.605} & \textbf{0.636} \\ \hline
	\end{tabular}
\vspace{-3mm}  %调整图片与上文的垂直距离
\end{table*}

\begin{figure}[h]
	\vspace{-0mm}  %调整图片与上文的垂直距离
	\setlength{\abovecaptionskip}{-0mm}   %调整图片标题与图距离
	\setlength{\belowcaptionskip}{-5mm}   %调整图片标题与下文距离
	\centering
	\vspace{-1.5mm}  %调整图片与上文的垂直距离
	\subfloat{\includegraphics[scale = 0.33]{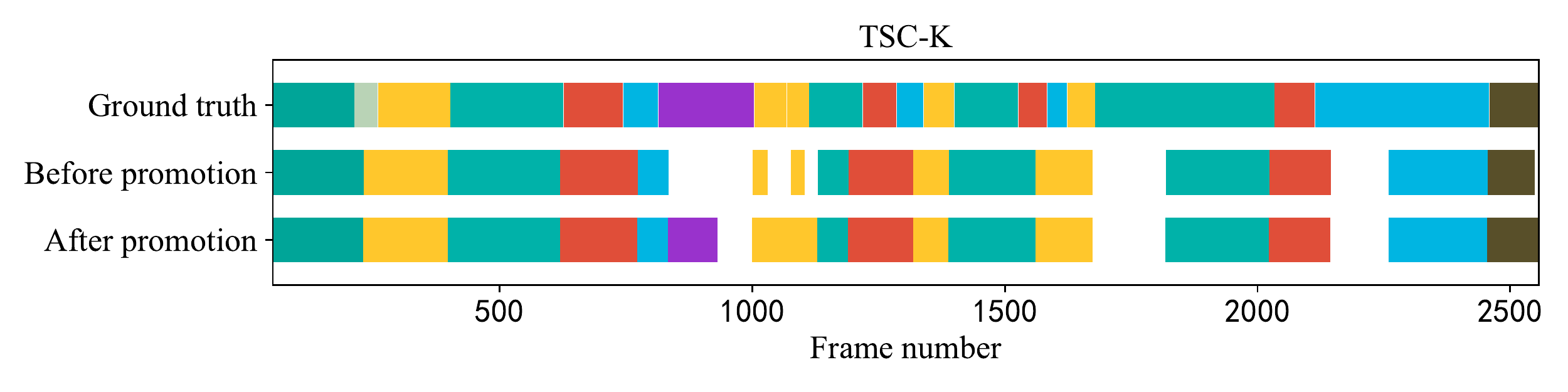}}
	\vspace{-1.5mm}  %调整图片与上文的垂直距离
	\subfloat{\includegraphics[scale = 0.33]{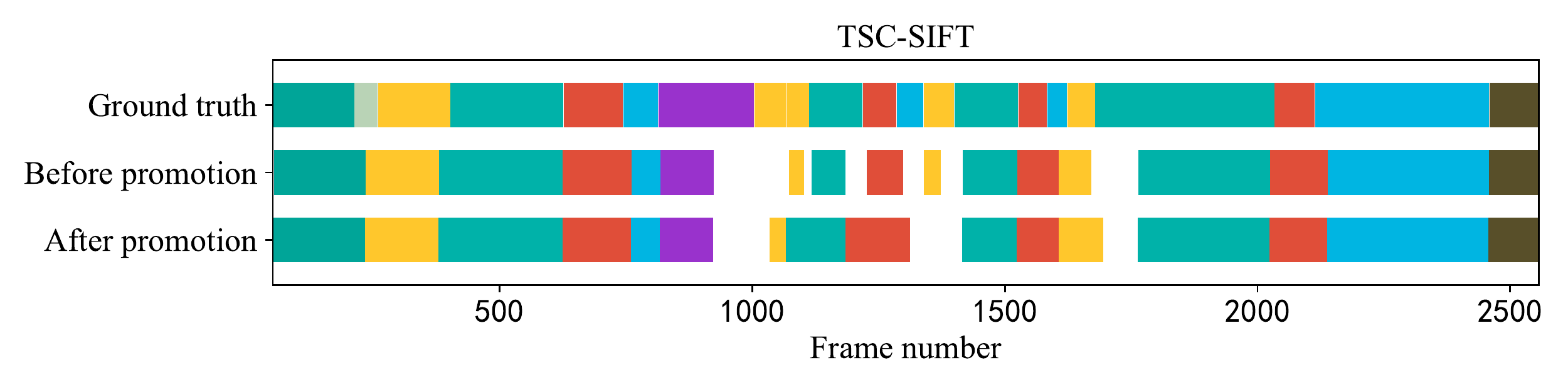}}
	\vspace{-1.5mm}  %调整图片与上文的垂直距离
	\subfloat{\includegraphics[scale = 0.33]{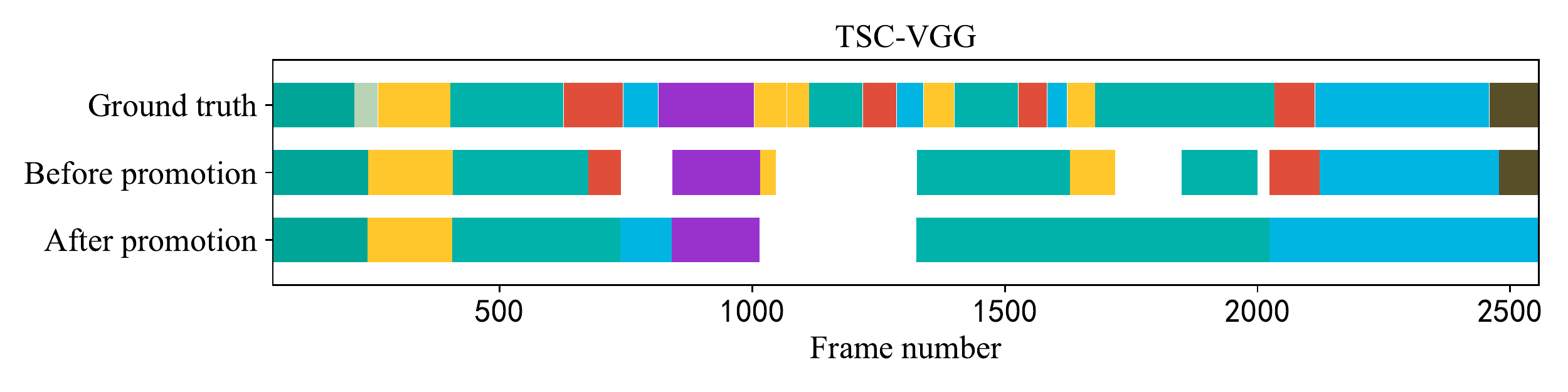}}
	\vspace{-1.5mm}  %调整图片与上文的垂直距离
	\subfloat{\includegraphics[scale = 0.33]{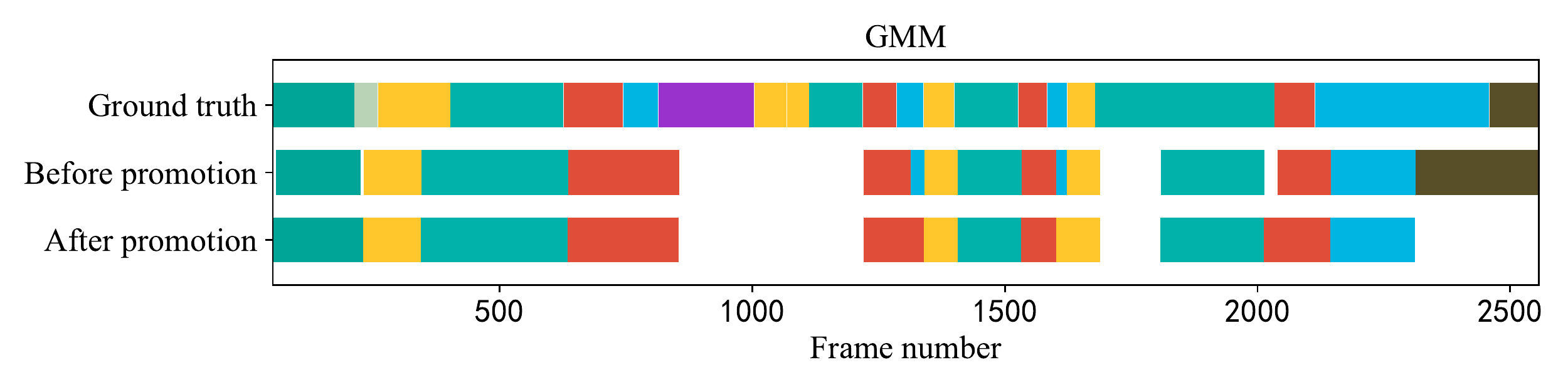}}
	\vspace{-1.5mm}  %调整图片与上文的垂直距离
	\subfloat{\includegraphics[scale = 0.33]{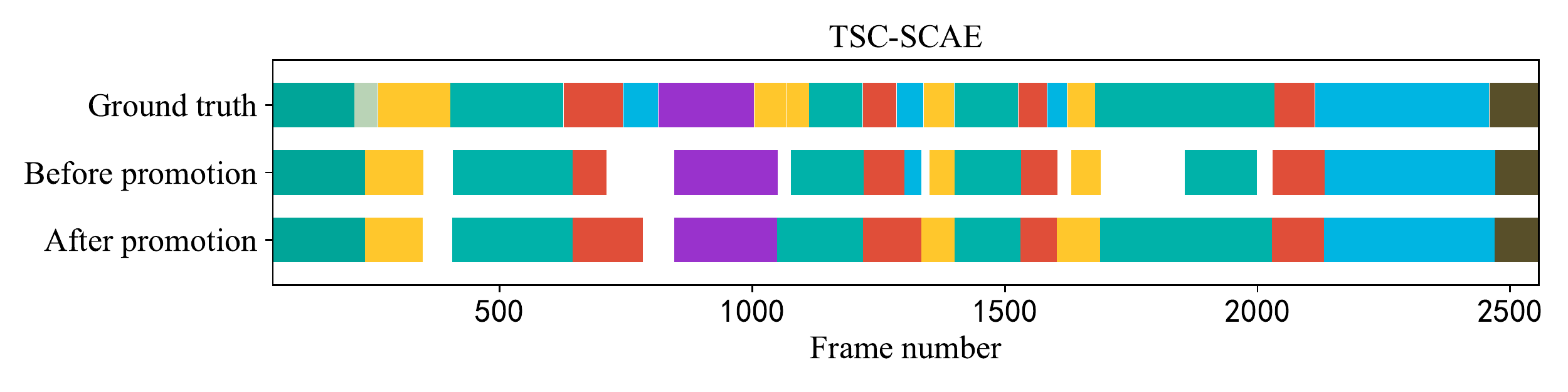}}
	\caption{Visualization of comparison of needle passing task.}
	\label{fig:segresult}
\end{figure}

As shown in TABLE \ref{tab:segacc}, the seg-acc of each method has been improved obviously for most cases. TSC-K is the biggest beneficiary with the improvement of seg-acc by 15.2$\%$ on average, while the accuracy is improved less for TSC-SIFT and TSC-VGG. In the experiment, we notice that it is difficult to refine the segmentation if the clustering results is far away from the ground truth. As shown in Fig. \ref{fig:segresult}, each color represents a surgical activity segment, while the white segment indicates incorrect segment or over-segmented segment. Among all methods, the seg-acc of GMM based method even declines after the promoting. Because GMM needs to specify the number of merged class, so over-segmentation in GMM is not very common instead is wrong segmentation. For our method TSC-SCAE, the segmentation promoting yields up to 16.7$\%$ improvement with respect to seg-acc. In most cases, the resultant segmentation after the promoting is significantly improved. From the view of TABLE \ref{tab:segacc}, we notice that the improvement of non-expert demonstration is more outstanding than the expert do, because the non-expert demonstration produces more over-segmentation fragments.
 
% The more similar the segmentation result and the ground truth are, the more accurate the segmentation result is.

 In  all experiments, TSC-SCAE obtains the best result of segmentation, it is proved that the proposed promoting method is very effective for the surgical trajectory segmentation. In general, it can be extended to most clustering segmentation algorithms.

\section{CONCLUSION}\label{sec:conclu}

This paper proposed a fast unsupervised method for surgical trajectory segmentation based on a compact stacking convolutional auto-encoder model and wavelet transform based filtering using multi-modal surgical demonstrations. The improvement with respect to the efficiency of segmentation is three-fold. First, new involved model can generate more discriminative visual features faster. Second, the short-range noises in the visual and kinematic features are filtered based on wavelet transform. Last but not least, a promoting approach is proposed to handle the over-segmentation problem. Compared with the state-of-the-art methods, experimental results demonstrate that the proposed algorithm can improve the accuracy of segmentation in an more efficient way.

\addtolength{\textheight}{-9cm}  % This command serves to balance the column lengths

%\section*{APPENDIX}


\section*{ACKNOWLEDGMENT}

This work was supported by the Project of Beijing Municipal Commission of Education (KM201710028017), National Natural Science Foundation of China (61702348, 61772351, 61602324), National Key R \& D Program of China (2017YFB1303000, 2017YFB1302800), the Project of the Beijing Municipal Science \& Technology Commission (LJ201607), Capacity Building for Sci-Tech Innovation - Fundamental Scientific Research Funds (025185305000), and Youth Innovative Research Team of Capital Normal University.

\bibliographystyle{IEEEtran}%unsrt,plain,splncs
\bibliography{surgical_segmentation}

\end{document}